\documentclass[sigconf]{acmart}

\usepackage[utf8]{inputenc}
\usepackage[english,vietnamese]{babel}
\usepackage{graphicx}
\usepackage{multirow}
\usepackage{booktabs}
\usepackage{enumitem}

\AtBeginDocument{%
  }

\acmConference[BDSIC]{The 2025 7th International Conference Big-data Service and Intelligent Computation
}{October 29--31,
  2025}{Bangkok, Thailand}

\begin{document}

\title{A Vietnamese Dataset for Text Segmentation and Multiple Choices Reading Comprehension}

\author{Toan Nguyen Hai}
\email{nguyenhaitoan@vnu.edu.vn}
\author{Ha Nguyen Viet}
\email{hanv@vnu.edu.vn}
\affiliation{%
  \institution{Institute for Artificial Intelligence, VNU University of Engineering and Technology}
  \city{Hanoi}
  \country{Vietnam}}

\author{Truong Quan Xuan}
\email{quanxuantruong2004@gmail.com}
\author{Duc Do Minh}
\email{dominhduc@vnu.edu.vn}
\affiliation{%
  \institution{Faculty of Information Technology, VNU University of Engineering and Technology}
  \city{Hanoi}
  \country{Vietnam}}

\renewcommand{\abstractname}{Abstract}
\begin{abstract}
Vietnamese, the 20th most spoken language with over 102 million native speakers, lacks robust resources for key natural language processing tasks such as text segmentation and machine reading comprehension (MRC). To address this gap, we present VSMRC, the \textbf{V}ietnamese Text \textbf{S}egmentation and \textbf{M}ultiple-Choice \textbf{R}eading \textbf{C}omprehension Dataset. Sourced from Vietnamese Wikipedia, our dataset includes 15,942 documents for text segmentation and 16,347 synthetic multiple-choice question-answer pairs generated with human quality assurance, ensuring a reliable and diverse resource. Experiments show that mBERT consistently outperforms monolingual models on both tasks, achieving an accuracy of 88.01\% on MRC test set and an F1 score of 63.15\% on text segmentation test set. Our analysis reveals that multilingual models excel in NLP tasks for Vietnamese, suggesting potential applications to other under-resourced languages. VSMRC is available at HuggingFace\footnote{\url{https://huggingface.co/VSMRC}}.
\end{abstract}

\maketitle



\section{Introduction}
\begin{figure*}[t!]
    \centering
    \includegraphics[width=0.7\textwidth]{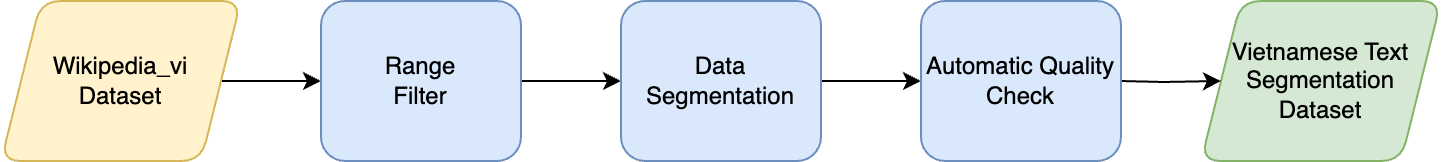}
    \caption{Illustration of the text segmentation component creation process starting from the Wikipedia\_vi Hugging Face Dataset.}
    \label{fig:Segmentation}
\end{figure*}

Machine reading comprehension (MRC) has emerged as a fundamental task in natural language processing, enabling computers to understand written text and answer questions about related content \cite{rajpurkar2016squad, conneau2023belebele, nguyen2020uitviquad}. This capability is essential for various applications including search engines, question-answering systems, and educational technology platforms \citep{joshi2017triviaqa, reddy2019coqa}. Similarly, text segmentation divides documents into coherent segments, each addressing a distinct topic or subtopic, facilitating tasks like information extraction \citep{prince2007text, shtekh2018exploring}, document summarization \citep{xiao2019extractive, liu2022end}. Moreover, recent research shows that effective text segmentation can significantly improve retrieval-augmented generation (RAG) systems, which combine retrieval and generation to produce accurate responses in modern NLP applications \citep{nguyen2024enhancing}.

For the Vietnamese language, spoken by over 102 million people worldwide, the development of robust NLP resources remains critical for advancing language technology applications. There has been significant progress in word segmentation \citep{kudod2025, nguyen2006wordseg, datquocnguyen}, essential for preprocessing due to Vietnamese's word boundary challenges. However, document-level topic segmentation, which organizes texts into coherent segments, remains underexplored. Similarly, general MRC tasks lack comprehensive datasets, as existing benchmarks like ViMMRC 2.0 \cite{luu2021multiple} for multiple-choice and UIT-ViNewsQA \cite{nguyen2022uit} for extractive question answering are task-specific and not broadly applicable across diverse comprehension scenarios.

To address these issues, we developed VSMRC, a new dataset for Vietnamese text segmentation and multiple-choice reading comprehension.The dataset addresses this research gap by providing a unified resource for both tasks. Unlike existing Vietnamese datasets that focus on individual tasks, VSMRC offers an evaluation framework where both tasks are designed to work independently while maintaining high standards for data quality and linguistic appropriateness. In this study, our main contributions of this paper are:

\begin{itemize}
    \item We present VSMRC, a comprehensive Vietnamese dataset containing over 15,000 documents for text segmentation and 16,000 question-answer pairs for multiple-choice task.
    
    \item A detailed analysis of question types, passage characteristics, and model performance provides valuable insights to guide future research in Vietnamese text understanding.
    
    \item Our work presents the first evaluation of open-domain text segmentation and multiple-choice reading comprehension for Vietnamese. Experiments with monolingual (English and Vietnamese) and multilingual models establish baselines and demonstrate VSMRC's utility for NLP research.
    
    \item Superior performance of multilingual models over monolingual models highlights their excellence for NLP studies across languages, proposing VSMRC could be use as a reosurce for cross-lingual study.
\end{itemize}

\section{Background}
\begin{figure*}[t!]
    \centering
    \includegraphics[width=\textwidth]{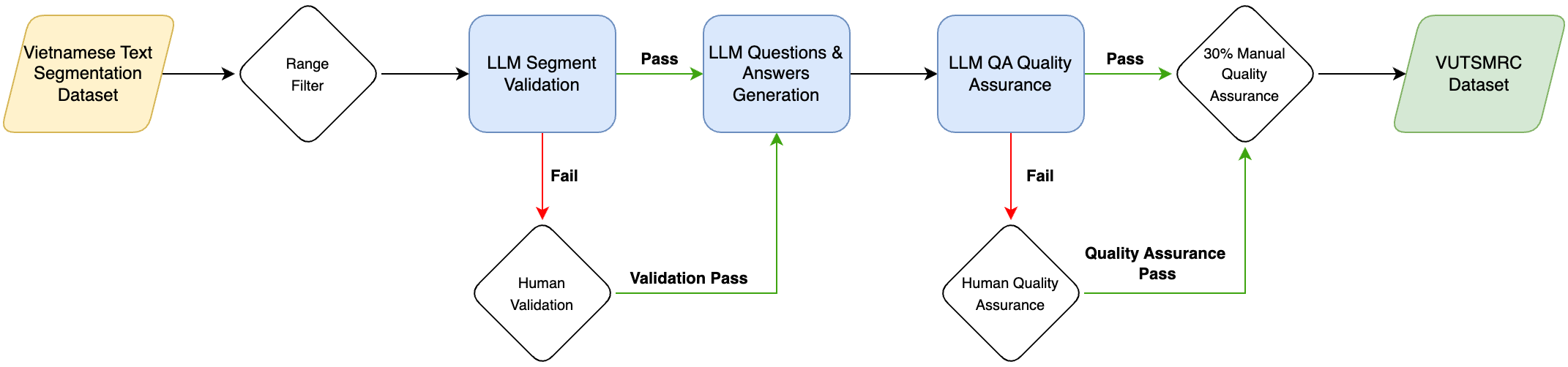}
    \caption{Illustration of the multiple-choice component creation process starting from the Vietnamese Text Segmentation Dataset.}
    \label{fig:MMRC}
\end{figure*}

\subsection{Text Segmentation Datasets}

Text segmentation involves dividing a document into coherent segments, each focusing on a distinct subtopic, which is crucial for applications like information extraction and document summarization. Several datasets support this task in high-resource languages. One of the largest is the \textbf{Wiki727k} dataset \citep{koshorek2018text}, which contains 727,000 English Wikipedia articles segmented into topics based on their tables of contents, widely used for supervised learning of text segmentation models. The \textbf{WikiSection} dataset \citep{arnold2019sector} contains 38,000 English and German Wikipedia articles labeled with 242,000 sections across disease and city domains, providing a rich resource for topic segmentation and classification. On the other hand, \textbf{TextSeg} \citep{xu2021rethinking} dataset focuses on scene and design text segmentation, providing annotations for either word or character-level segmentation in images, though it is less relevant for topical segmentation. The \textbf{Choi} dataset \citep{choi2000advances}, a smaller but standard benchmark, includes concatenated texts for evaluating segmentation algorithms. 

\subsection{Machine Reading Comprehension Datasets}

Machine Reading Comprehension requires a model to read a passage and answer multiple-choice questions, testing its comprehension and reasoning abilities. In English, prominent datasets include \textbf{SQuAD} \citep{rajpurkar2016squad}, which provides over 100,000 question-answer pairs based on Wikipedia articles, \textbf{RACE} \citep{lai2017race}, a large-scale dataset collected from English exams for Chinese students containing multiple-choice questions that demand deep understanding, and \textbf{Cosmos QA} \citep{huang2019cosmos}, which focuses on commonsense reasoning over everyday narratives, also using a multiple-choice format. The multilingual \textbf{MLQA} dataset \citep{lewis2020mlqa} includes 12,738 questions and answers in English, and 5,029 in other languages including Vietnamese. Another multilingual dataset is \textbf{BELEBELE} \citep{conneau2023belebele}, which covers 122 language variants including Vietnamese, provides 900 questions. However, neither of these datasets is specifically tailored to Vietnamese linguistic needs and cultural context.

\subsection{Vietnamese NLP Datasets}

Vietnamese NLP faces challenges due to limited datasets. For machine reading comprehension, \textbf{UIT-ViQuAD} \citep{nguyen2020uitviquad} contains over 23,000 question-answer pairs from Vietnamese Wikipedia articles, but it focuses on span-extraction, not multiple-choice formats. Similarly, \textbf{UIT-ViNewsQA} \citep{nguyen2022uit} targets news articles with the same span-extraction approach, while \textbf{ViCoQA} \citep{luu2021conversational} introduces conversational MRC with 2,000 conversations and 10,000 question-answer pairs from health news. For multiple-choice reading comprehension specifically, \textbf{ViMMRC 2.0} \citep{luu2021multiple} is the only dedicated dataset but remains constrained by its limited size and domain scope. This highlights a significant lack of general-purpose multiple-choice reading comprehension resources for Vietnamese.

For text segmentation, no dedicated datasets exist in Vietnamese prior to our work, despite progress in lower-level tasks like word segmentation through tools such as \textbf{VnCoreNLP} \citep{vu2018vncorenlp} and \textbf{Underthesea} \footnote{\url{https://undertheseanlp.com/}} . By providing resources for both text segmentation and general multiple-choice reading comprehension, VSMRC fills these critical gaps and enables research into Vietnamese NLP across these important tasks.
\section{The VSMRC Dataset}

In this section, we introduce VSMRC, which follows a two-phase approach: first, we develop the text segmentation component using Vietnamese Wikipedia articles; then, we leverage these segmented texts to generate synthetic multiple-choice questions for machine reading comprehension evaluation. This dual-component design enables both text segmentation research and reading comprehension assessment within a single unified resource. Our methodologies for each phase are described below:

\subsection{Text Segmentation Data Creation}

We create the text segmentation component using the Vietnamese Wikipedia dataset (version 20250301) from Hugging Face \footnote {\url{https://huggingface.co/datasets/vietgpt/wikipedia_vi}} as our starting point. Figure \ref{fig:Segmentation} illustrates this creation pipeline.


\subsubsection{\textbf{Data Acquisition and Initial Filtering}}
We filter articles to include only those with 750-3,000 tokens, specifically targeting articles that average approximately 1,500 tokens. This range aligns with established benchmarks like Wiki727k and Wikisection \cite{yu2023improving}. We then extract article URLs for crawler-based content retrieval, accounting for potential Wikipedia updates.

\subsubsection{\textbf{Structure Validation and Segmentation}}
We retrieve articles via URL crawling and segment them into headers and content sections. All formatting elements—including HTML markup, wiki tags, lists, and tables—are removed to produce clean plain text.

\subsubsection{\textbf{Quality Assurance and Final Filtering}}
For the last step, we perform these post-processing steps:

\begin{itemize}
    \item Remove segments with only one sentence, as they don't help with learning segment boundaries.
    
    \item Remove documents with fewer than three segments, as they don't have enough section changes.
    
    \item Remove documents where more than 80\% of segments were altered.
    
    \item Verify that all articles remained within our target length after filtering
\end{itemize}

\subsection{Multiple Choice Data Creation}
\label{sec:multiple_choice_data_creation}
Building upon the previous component, we develop a synthetic multiple-choice reading comprehension component for VSMRC. Our methodology combines large language models (LLMs): Google Gemini 2.0 Flash Lite\footnote{\url{https://cloud.google.com/vertex-ai/generative-ai/docs/models/gemini/2-0-flash-lite}}, OpenAI GPT-4o-mini\footnote{\url{https://openai.com/index/gpt-4o-mini-advancing-cost-efficient-intelligence/}}, and DeepSeek-V3\footnote{\url{https://api-docs.deepseek.com/news/news1226}} for our QA generation and validation. We also leverage human expertise, specifically a team of 20 experienced experts in Vietnamese literature, throughout the validation process to ensure dataset quality and reliability. Figure \ref{fig:MMRC} demonstrates our multiple-choice component creation pipeline.

\begin{table}[b!]
\centering
\caption{Types of questions in the VSMRC dataset.}
\label{tab:question_types}
\resizebox{\columnwidth}{!}{
\begin{tabular}{c|p{7cm}}
\textbf{Type} & \textbf{Description} \\
\hline
\textbf{Reasoning} & {
Questions that require synthesizing multiple sentences to explain a concept, cause, or significance.
} \\
\hline
\textbf{Fact-check} & {
Questions that test specific details, events, or relationships with one or two sentences .
} \\
\hline
\textbf{Fill-blank} & {
Questions where the reader must complete a statement with missing words or phases.
} \\
\hline
\textbf{List} & {
Questions that ask to identify a correct group or sequence of items from multiple details across the paragraph.
} \\
\hline
\end{tabular}
}
\end{table}

\subsubsection{\textbf{Segment Range Filter}}
We filter segments to keep only those within the 450--1,200 characters, matching the segment size in many MRC datasets such as SQuAD \cite{rajpurkar2016squad} and UIT-ViQuAD \cite{nguyen2020uitviquad}. This ensures segments are within the suitable range for creating meaningful QA pairs.

\subsubsection{\textbf{Segment Quality Validation}}
We validate segments using OpenAI and Gemini to ensure quality. The validation prompt (Appendix \ref{app:validation_prompt}) checks each document for segments that contain sensitive content, accuracy issues, lack of specificity, duplicates, clarity problems, poor distractor potential, or inappropriate complexity. Segments rejected by either model are reviewed by three text reading experts using the same criteria to confirm suitability for QA generation.

\subsubsection{\textbf{Multiple Choice Questions and Answers Generation}}
We employ Gemini 2.0 Flash Lite to generate a single multiple-choice question per qualified segment, with four answer choices including one correct response. The generation prompt (Appendix \ref{app:generation_prompt}) ensures questions maintain passage-dependence, cultural appropriateness, and coverage across all four question types. All distractors derive directly from the passage content to prevent answer bias and maintain question validity. Table~\ref{tab:question_types} provides specifications for each question type included in the MRC component, and appendix \ref{app:detail_questions_type} shows more detail how each questions looks like. 

\subsubsection{\textbf{LLMs Quality Assurance}}
Generated QA pairs undergo verification by two LLMs, DeepSeek-V3 and OpenAI GPT-4o-mini, using a verification prompt (Appendix \ref{app:verification_prompt}). The validation process evaluates question clarity, answer accuracy, distractor quality, and content appropriateness. When either model flags a QA pair as unsuitable, three text reading experts conduct a review using identical criteria to confirm the final assessment.

\begin{table}[b!]
\centering
\caption{Overall experts's quality assurance result.}
\label{tab:qual}
\begin{tabular}{lc}
\toprule
\textbf{Criterion} & \textbf{Percentage (\%)} \\
\midrule
Low quality distractors & 4.58 \\
Unsuitable Question & 3.06 \\
Unclear answer & 7.6 \\
Outside knowledge & 3.65 \\
\bottomrule
\end{tabular}
\end{table}

\subsubsection{\textbf{Human Quality Assurance}}
\label{sec:human_quality_assurance}
To ensure rigorous quality control, each of the 20 experts evaluated 300 randomly selected QA pairs, totaling 6,000 pairs (37\% of the MRC components). Experts assess each QA pair using the following validation criteria:
\begin{itemize}[leftmargin=*]
\item \textbf{Low quality distractors}: Distractors are flagged if they are unrelated to the passage or too easily eliminated, making the correct answer overly obvious. For example, distractors that do not derive from the passage content or are significantly different in length or context from the correct answer.
\item \textbf{Unsuitable question}: Questions are flagged if they are irrelevant to the passage or misrepresent its content. For example, a question addressing a topic not covered in the passage.
\item \textbf{Unclear answer}: Questions are flagged if they have multiple correct answers or if the correct answer is ambiguous based on the passage. For example, a vague question or one where the passage lacks sufficient information to identify a single correct answer.
\item \textbf{Outside knowledge}: Questions are flagged if answering them needs information beyond the passage. For example, a question necessitating historical or scientific information not mentioned in the passage.
\end{itemize}
Table \ref{tab:qual} indicates that only a small percentage of QA pairs contains issues, with less than 10\% containing some issues. These low error rates validate the effectiveness of our multi-stage quality assurance process and demonstrate the suitability of the vast majority of QA pairs for inclusion in the VSMRC dataset.


\section{Data Analysis}

\subsection{\textbf{Overall Statistics}}
The statistics of the VSMRC dataset are presented in Table \ref{tab:dataset_stats}, showing the distribution across training (Train), development (Dev), and test (Test) sets. The dataset consists of 15,942 documents for text segmentation task with 15,942 documents and a multiple-choice component containing 16,347 questions. For text segmentation, we observed an average of 6.38 segments per document, with each document containing approximately 47.26 sentences and 1,226 tokens on average. For multiple-choice, questions had an average length of 14.50 words, while answers are notably concise with an average of 6.44 words. The passages from which questions are derived have an average length of 159.20 words.

\begin{table}[t!]
\centering
\caption{Overview statistics of the VSMRC dataset: text segmentation (top) and multiple choice (bottom)}
\begin{tabular}{lrrrr}
\hline
\multicolumn{5}{c}{\textbf{Text Segmentation Component}} \\
\hline
\multicolumn{1}{c}{\textbf{}} & \multicolumn{1}{c}{\textbf{Train}} & \multicolumn{1}{c}{\textbf{Dev}} & \multicolumn{1}{c}{\textbf{Test}} & \multicolumn{1}{c}{\textbf{All}} \\ \hline
Num Documents   & 12,752            & 1,594     & 1,596         & {\bf 15,942} \\ 
Avg segments/doc   & 7.04        & 6.06      & 6.04        & {\bf 6.38}    \\ 
Avg sentences/doc  & 44.76        & 48.47     & 48.56       & {\bf 47.26}   \\ 
Avg tokens/doc     & 1,229        & 1,217     & 1,231       & {\bf 1,226}   \\ 
\hline
\multicolumn{5}{c}{\textbf{Multiple Choice Component}} \\
\hline
\multicolumn{1}{c}{\textbf{}} & \multicolumn{1}{c}{\textbf{Train}} & \multicolumn{1}{c}{\textbf{Dev}} & \multicolumn{1}{c}{\textbf{Test}} & \multicolumn{1}{c}{\textbf{All}} \\ \hline
Num Questions   & 11,442         & 3,269   & 1,636         & {\bf 16,347} \\ 
Num Fact Check Questions   & 6,015        & 1,678   & 859         & {\bf 8,552}  \\ 
Num Fill Blank Questions   & 221          & 62      & 30          & {\bf 313}    \\ 
Num Reasoning Questions    & 4,639        & 1,371   & 686         & {\bf 6,696}  \\ 
Num List Questions         & 524          & 152     & 58          & {\bf 734}    \\ 
Avg question length   & 14.52        & 14.43   & 14.56       & {\bf 14.5}   \\ 
Avg answer length   & 6.49         & 6.51    & 6.20        & {\bf 6.44}   \\ 
Avg passage length & 159.08     & 159.20  & 159.73      & {\bf 159.2}  \\ 
\hline
\end{tabular}
\label{tab:dataset_stats}
\end{table}

\subsection{\textbf{Multiple Choice Component Length Statistics}}

We analyzed our multiple-choice component according to three categories of length: question length, answer length (both in Table \ref{tab:qalength}), and passage length (in Table \ref{tab:readingtextlength}). Questions with 11-15 words dominated the dataset, accounting for 43.93\% of all questions. For answers, shorter responses are clearly prevalent, with 1-5 word answers constituting 52.33\% of the dataset, and 6-10 word answers accounting for 28.03\%. Regarding passage length, the dataset showed a balanced distribution between the 101-150 word range (42.60\%) and the 151-200 word range (43.67\%), together comprising 86.27\% of all passages. These length distributions highlight the distinctive characteristics of our Vietnamese multiple-choice reading comprehension component.

\begin{table}[t!]
\centering
\caption{Statistics of question and answer lengths in our multiple choice component}
\begin{tabular}{crrrr}
\hline
\multirow{2}{*}{\textbf{Length}} & \multicolumn{4}{c}{\textbf{Question}} \\ \cline{2-5} 
 & \multicolumn{1}{c}{\textbf{Train}} & \multicolumn{1}{c}{\textbf{Dev}} & \multicolumn{1}{c}{\textbf{Test}} & \multicolumn{1}{c}{\textbf{All}} \\ \hline
1-5 & 0.6 & 0.6 & 0.7 & 0.63 \\ 
1-10 & 19.5 & 21.4 & 17.7 & 19.53 \\ 
11-15 & {\bf 43.9} & {\bf 42.4} & {\bf 45.5} & {\bf 43.93} \\ 
16-20 & 25.6 & 24.4 & 24.6 & 24.87 \\ 
\textgreater{}20 & 10.4 & 11.2 & 11.5 & 11.04 \\ \hline
\multirow{2}{*}{\textbf{Length}} & \multicolumn{4}{c}{\textbf{Answer}} \\ \cline{2-5} 
 & \multicolumn{1}{c}{\textbf{Train}} & \multicolumn{1}{c}{\textbf{Dev}} & \multicolumn{1}{c}{\textbf{Test}} & \multicolumn{1}{c}{\textbf{All}} \\ \hline
1-5 & {\bf 52.3} & {\bf 51.7} & {\bf 53.0} & {\bf 52.33} \\ 
6-10 & 27.6 & 27.7 & 28.8 & 28.03 \\ 
11-15 & 13.0 & 13.7 & 11.7 & 12.80 \\ 
16-20 & 4.9 & 5.1 & 5.0 & 5.00 \\ 
\textgreater{}20 & 2.2 & 1.8 & 1.5 & 1.84 \\ \hline
\end{tabular}
\label{tab:qalength}
\end{table}

\begin{table}[t]
\centering
\caption{Statistics of passage lengths in our multiple choice component}
\begin{tabular}{crrrr}
\hline
\multirow{2}{*}{\textbf{Length}} & \multicolumn{4}{c}{\textbf{Passage}} \\ \cline{2-5} 
 & \multicolumn{1}{c}{\textbf{Train}} & \multicolumn{1}{c}{\textbf{Dev}} & \multicolumn{1}{c}{\textbf{Test}} & \multicolumn{1}{c}{\textbf{All}} \\ \hline
\textless{}101 & 0.6 & 0.7 & 0.4 & 0.57 \\ 
101-150 & {\bf 43.3} & 42.6 & 41.9 & 42.60 \\ 
151-200 & 43.0 & {\bf 43.3} & {\bf 44.7} & {\bf 43.67} \\ 
\textgreater{}201 & 13.1 & 13.4 & 13.0 & 13.16 \\ \hline
\end{tabular}
\label{tab:readingtextlength}
\end{table}
\subsection{\textbf{Type Based Analysis}}

As part of the human quality assurance process in Section \ref{sec:human_quality_assurance}, experts were also required to analyzed reasoning types for each synthetic multiple-choice question. Following Nguyen et al. \cite{nguyen2020uitviquad}, questions were manually categorized into five types of reasoning: word matching (WM), paraphrasing (PP), single-sentence reasoning (SSR), multi-sentence reasoning (MSR), and ambiguous or insufficient (AoI).
Table \ref{tab:question_types_sentences} presents the distribution of reasoning types across different question categories based on the expert annotation. Our analysis revealed distinct reasoning patterns associated with each question type.

Fact-check questions primarily relied on word matching (34.48\%) and single-sentence reasoning (24.46\%), which aligns with their purpose of testing specific details from the text. 

Fill-blank questions showed an almost equal distribution between single-sentence reasoning (39.52\%) and word matching (39.49\%), suggesting these questions typically require understanding the immediate context around the blank.

Reasoning questions, as expected, predominantly required multi-sentence reasoning (40.62\%) and paraphrasing (29.49\%). The minimal reliance on word matching (5.93\%) confirms that these questions test deeper comprehension abilities, requiring readers to synthesize information across multiple sentences and make inferences beyond what is explicitly stated.

List questions demonstrated a balanced distribution across paraphrasing (33.16\%), multi-sentence reasoning (32.01\%), and single-sentence reasoning (28.08\%). This suggests that identifying correct groups or sequences often requires comprehending relationships between different parts of the passage.

These patterns confirm that our automated generation system produces questions aligned with their intended reasoning demands, ensuring comprehensive evaluation of reading comprehension at various cognitive complexity levels.

\begin{table}[!t]
\centering
\caption{Distribution of reasoning strategies across question types in VSMRC based on expert annotation}
\begin{tabular}{lccccc}
\toprule
\textbf{Type} & \textbf{SSR (\%)} & \textbf{WM (\%)} & \textbf{PP (\%)} & \textbf{MSR (\%)} & \textbf{AoI (\%)} \\
\midrule
fact\_check & 24.46 & \textbf{34.48} & 11.49 & 22.52 & 7.05 \\
fill\_blank & \textbf{39.52} & 39.49 & 9.66 & 8.91 & 2.42 \\
reasoning & 17.83 & 5.93 & 29.49 & \textbf{40.62} & 6.13 \\
list & 28.08 & 1.90 & \textbf{33.16} & 32.01 & 4.85 \\
\bottomrule
\end{tabular}
\label{tab:question_types_sentences}
\end{table}
\section{Experiments}
\label{sec:experiments}

\subsection{Evaluation Models}
\subsubsection{\textbf{Text Segmentation}}
As this study presents the first Vietnamese text segmentation evaluation, no established baselines currently exist. To provide a robust initial benchmark for this novel task, we evaluated a diverse set of prominent models. We categorized them as follows:
\begin{itemize}
    \item \textbf{Vietnamese Monolingual Models:} Following the principle that BERT and its variants serve as suitable standard baselines for many NLP tasks \cite{rogers2020primer}, we included models pre-trained on large Vietnamese corpora such as: PhoBERT-large \cite{nguyen2020phobert} and ViDeBERTa-base \cite{tran2023videberta}, which have demonstrated good performance on various Vietnamese NLP benchmarks.
    \item \textbf{Multilingual Models:} We incorporated XLM-R-base \cite{conneau2020xlmr} and mBERT \cite{devlin2019bert}, as these models are widely recognized and frequently used baselines on both monolingual and multilingual datasets across diverse NLP tasks \cite{conneau2023belebele, nguyen2020uitviquad}.
    \item \textbf{English Monolingual Models:} We also included BERT-base-cased \cite{devlin2019bert} and Electra-base-discriminator \cite{clark2020electra}. Although pre-trained on English, these models represent highly influential architectures that have shown strong performance in English text segmentation \cite{yu2023improving, lo2021transformer}. Their inclusion provided valuable comparative insights into how powerful general-purpose English models perform on this Vietnamese task.
\end{itemize}

\subsubsection{\textbf{Multiple-Choice Reading Comprehension}}

Using baselines from the ViMMRC 2.0 study \cite{luu2021multiple}, we evaluated mBERT \cite{devlin2019bert}, XLM-R\textunderscore base \cite{conneau2020xlmr}, viBERT \cite{bui2020improving}, and BERT4News \cite{nguyen2021nlpbk}. Additionally, we included PhoBERT \cite{nguyen2020phobert} for broader comparison.

\subsection{Experimental Settings}
Both tasks are trained with a learning rate of 2e-05, batch size of 4 for text segmentation (and 8 for MRC), and 3 epochs, using the AdamW optimizer \cite{loshchilov2018adamw}. The text segmentation component was split into 80\% training, 10\% development, and 10\% test sets, while the multiple-choice component was split into 70\% training, 20\% development, and 10\% test sets. 

\subsection{Task-Specific Details and Metrics}

\subsubsection{\textbf{Text Segmentation}}
\label{subsec:text_segmentation} 
We evaluated text segmentation using three metrics: \textbf{F1}, \textbf{P$_k$} \cite{beeferman1999pk}, and \textbf{WindowDiff (WD)} \cite{pevzner2002windowdiff}. F1 measures boundary detection accuracy through precision and recall. P$_k$ addresses F1's limitations by measuring segmentation errors within a sliding window. WD improves on P$_k$ by better handling varying segment sizes and near-miss predictions. For both P$_k$ and WD, we used a window size of half the average segment length. Higher F1 scores and lower P$_k$ and WD values indicate better performance. 

\subsubsection{\textbf{Multiple-Choice MRC}}
\label{subsec:mrc}
We evaluated the MRC task using \textbf{Accuracy}, which measures the percentage of questions where the model selects the correct answer choice.
\section{Results}
\subsection{Text Segmentation}
\begin{table}[b!]
    \centering
    \caption{Performance of text segmentation baselines on our dataset}
    \label{tab:nq_metrics_result}
    \small
    \begin{tabular}{l|ccc}
        \hline
        \textbf{Model} & \textbf{F$_1\uparrow$} & \textbf{$P_{k}\downarrow$} & \textbf{$W\!D\downarrow$} \\
        \hline
        \multicolumn{4}{c}{\textbf{Dev Set}} \\
        \hline
        PhoBERT\_large & 37.02 & 36.01 & 39.38 \\
        ViDeBerta\_base & 53.15 & 28.30 & 31.43 \\
        BERT\_base & 9.38 & 45.4 & 46.51 \\
        Electra\_base & 4.81 & 42.29 & 42.76 \\
        XLM\_R\_base & \textbf{63.33} & 23.52 & 26.79 \\
        mBERT & 63.08 & \textbf{23.43} & \textbf{26.70} \\
        \hline
        \multicolumn{4}{c}{\textbf{Test Set}} \\
        \hline
        PhoBERT\_large & 36.51 & 35.04 & 37.79 \\
        ViDeBerta\_base & 52.97 & 27.61 & 30.54 \\
        BERT\_base & 8.35 & 39.89 & 40.78 \\
        Electra\_base & 4.96 & 40.68 & 41.15 \\
        XLM\_R\_base & 62.83 & 23.40 & 26.52 \\
        mBERT & \textbf{63.15} & \textbf{23.16} & \textbf{26.28} \\
        \hline
    \end{tabular}
\end{table}

Table~\ref{tab:nq_metrics_result} presents the results for text segmentation performance. Surprisingly, multilingual models significantly outperformd both Vietnamese-specific and English monolingual models. On the test set, mBERT achieved the highest F1 score of 63.15\%, closely followed by XLM-R-base at 62.83\%. In contrast, other models performed substantially worse. Vietnamese models (PhoBERT\_large and ViDeBerta\_base) reached F1 scores up to 52.97\%, while English models (BERT\_base and Electra\_base) performed poorly, with F1 scores as low as 4.96\%. The expected underperformance of English models reflected language mismatch, while Vietnamese models' weaker results suggested that multilingual training captures broader language patterns advantageous for this task.
\subsubsection{\textbf{Effects of Multilingual Training Data on Text Segmentation}}

Given the superior performance of multilingual models (mBERT and XLM-R) over Vietnamese-specific models on our data, we hypothesized that multilingual training data might further improve results. To test this, we combined our text segmentation component with 15,000 random English documents from Wiki727k \cite{koshorek2018text} dataset and compared the performance with the Vietnamese-only training baseline.

\begin{table}[!t]
    \centering
    \caption{Performance of text segmentation models with Vietnamese-only vs. combined training data}
    \label{tab:multilingual_training_results}
    \small
    \begin{tabular}{l|c|ccc}
        \hline
        \textbf{Model} & \textbf{Training Data} & \textbf{F$_1\uparrow$} & \textbf{$P_{k}\downarrow$} & \textbf{$W\!D\downarrow$} \\
        \hline
        \multicolumn{5}{c}{\textbf{Dev Set}} \\
        \hline
        XLM\_R\_base & Vietnamese only & 63.33 & 23.52 & 26.79 \\
        mBERT & Vietnamese only & 63.08 & 23.43 & 26.70 \\
        XLM\_R\_base & Vietnamese + English & \textbf{64.63} & 22.94 & 26.10 \\
        mBERT & Vietnamese + English & 63.30 & \textbf{22.89} & \textbf{25.96} \\
        \hline
        \multicolumn{5}{c}{\textbf{Test Set}} \\
        \hline
        XLM\_R\_base & Vietnamese only & 62.83 & 23.40 & 26.52 \\
        mBERT & Vietnamese only & 63.15 & 23.16 & 26.28 \\
        XLM\_R\_base & Vietnamese + English & 63.72 & 22.88 & 25.83 \\
        mBERT & Vietnamese + English & \textbf{64.38} & \textbf{22.44} & \textbf{25.50} \\
        \hline
    \end{tabular}
\end{table}

Table~\ref{tab:multilingual_training_results} shows that both models performed better when trained on the combined dataset. For mBERT, the F1 score increased from 63.15\% to 64.38\% on the test set, while $P_k$ and WD errors decreased. XLM-R-base showed similar improvements, indicating that adding English training data improves Vietnamese text segmentation performance. The improvement might happen because:

\begin{itemize}
    \item Both languages may share some text structure patterns despite being from different language families.
    
    \item More training examples help the model identify common features of segment boundaries.
    
    \item The additional data prevents the model from focusing too much on Vietnamese-specific patterns.
    
    \item Multilingual models like mBERT and XLM-R can effectively use their pre-trained knowledge of multiple languages.
\end{itemize}

These findings suggest that using data from multiple languages could be a useful approach for improving Vietnamese NLP tasks in general.

\subsection{Multiple-Choice Reading Comprehension }

\begin{table}[t]
    \centering
    \caption{Performance of MRC models on our dataset}
    \label{tab:mrc_metrics}
    \small
    \begin{tabular}{l|cc}
        \hline
        \textbf{Model} & \textbf{Dev Acc}$\uparrow$ & \textbf{Test Acc}$\uparrow$ \\
        \hline
        mBERT & \textbf{88.68} & \textbf{88.01} \\
        XLM-R-base & 87.14 & 87.89 \\
        viBERT & 85.66 & 84.45 \\
        BERT4News & 54.51 & 54.15 \\
        PhoBERT-base & 84.25 & 83.77 \\
        \hline
    \end{tabular}
\end{table}

For our MRC's result outlined in Table~\ref{tab:mrc_metrics}, mBERT still achieved the highest development set accuracy at 88.68\%, but other Vietnamese monolingual models such as PhoBERT\_base also performed competitively at 84.25\%. This contrasts with the text segmentation results, where Vietnamese monolingual models significantly underperformed against multilingual models. The difference in relative performances suggests that the skills required for text segmentation and reading comprehension in Vietnamese are distinct, with language-specific pre-training providing more benefits for comprehension tasks. 

\subsubsection{\textbf{Effects of Type-Based Aspects on MRC}}

\begin{figure}[t]
    \centering
    \includegraphics[width=0.45\textwidth]{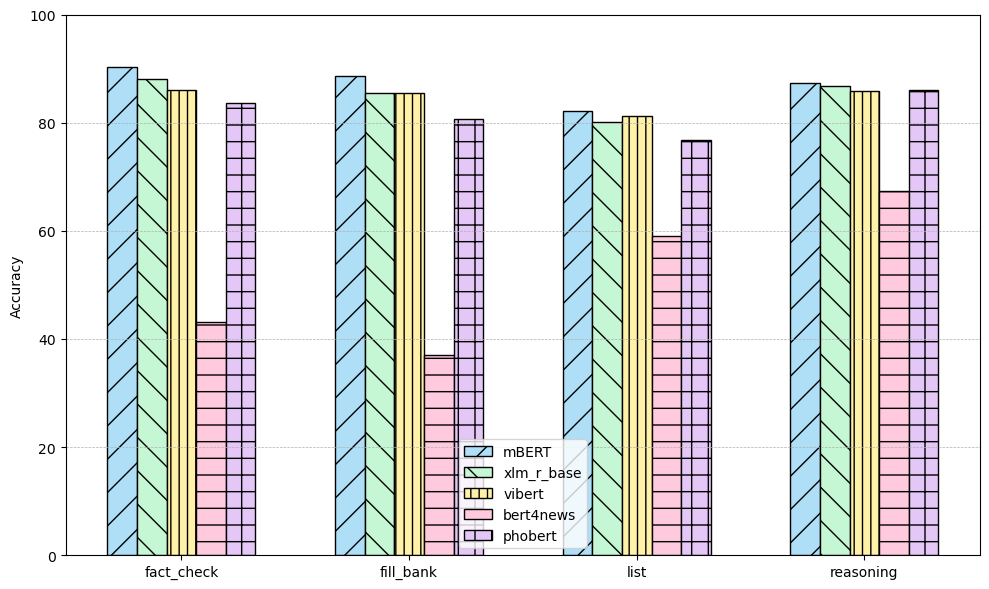}
    \caption{Type-based analysis of Accuracy performances on the VSMRC's MRC task test set.}
    \label{fig:question_type}
\end{figure}

To gain deeper insights into the performance of MRC models for Vietnamese, we analyzed their accuracy across different question types. Figure~\ref{fig:question_type} presents the type-based analysis of accuracy on the test set for the MRC component of VSMRC. The results showed that mBERT consistently outperformed other models across all question types, achieving the highest accuracy. In contrast, BERT4News consistently underperformed, with notably lower accuracies (e.g., 43.15\% for fact-check and 67.32\% for reasoning). Among the question types, complex types such as reasoning and list exhibited slightly lower performance compared to simpler types like fact-check and fill-blank. This trend aligns with the higher reasoning demands of reasoning and list questions, which often require multi-sentence reasoning or paraphrasing, as observed in similar datasets like UIT-ViQuAD~\cite{nguyen2020uitviquad}.

\begin{figure}[t]
    \centering
    \includegraphics[width=0.45\textwidth]{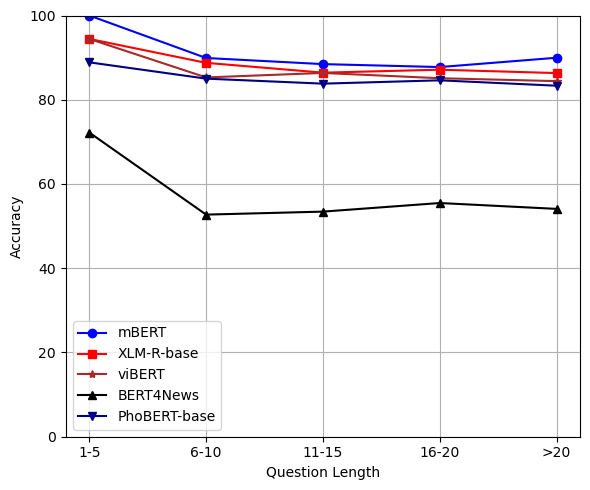}
    \caption{Accuracy of MRC models across question lengths on the VSMRC's multiple-choice test set.}
    \label{fig:question_length}
\end{figure}

\begin{figure}[t]
    \centering
    \includegraphics[width=0.45\textwidth]{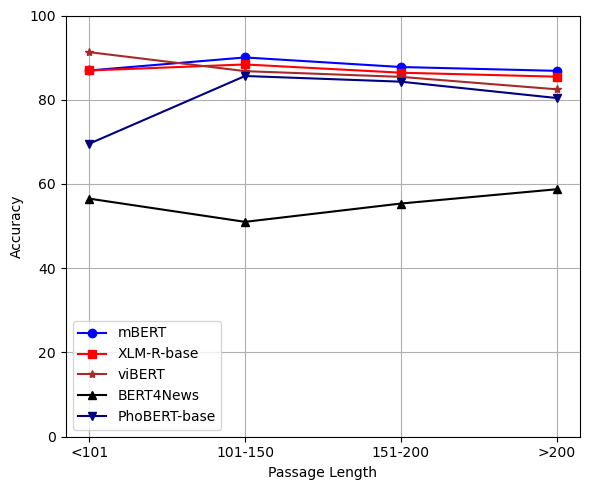}
    \caption{Accuracy of MRC models across passage lengths on the VSMRC's multiple-choice test set.}
    \label{fig:passage_length}
\end{figure}

\subsubsection{\textbf{Effects of Length-Based Aspects on MRC}}

For question length (Figure~\ref{fig:question_length}), we observed that performance tended to decline as questions became longer. Short questions (1-5 words) yielded high accuracies (100\% for mBERT, 94.44\% for viBERT), though these represent only 0.63\% of the dataset. The most common question length category (11-15 words, which comprises 43.93\% of questions) showed strong but slightly lower performance, with accuracies ranging from 83.82\% (PhoBERT-base) to 88.46\% (mBERT). For the longest questions (>20 words), performance further decreased across all models, suggesting that longer questions introduce additional complexity that challenges the models' comprehension capabilities.

Regarding passage length (Figure~\ref{fig:passage_length}), we observed that most models performed optimally with passages of moderate length. viBERT achieved its peak performance with short passages (<101 words) at 91.30\%, while mBERT and XLM-R-base excelled with passages of 101-150 words (90.04\% and 88.38\%, respectively). Notably, performance tended to decline for longer passages across most models.

\section{Conclusion}
Comprehensive resources for Vietnamese NLP tasks were essential for advancing language applications for this widely-spoken language. We addressed this need by creating VSMRC, a dataset supporting both text segmentation and multiple-choice reading comprehension. Empirical results highlight the superior performance of multilingual models over monolingual counterparts on text segmentation, while Vietnamese-specific models showed competitive results on MRC task, underscoring the task-specific advantages of language-focused pretraining.

Our analysis reveals several directions for model improvement. The performance decline observed with longer questions and passages suggests the need for models that can better handle lengthy contexts. Similarly, the varying performance across different question types points to the need for reasoning-focused architectures that can better handle complex inferential tasks, particularly for list and reasoning questions which proved more challenging.

Furthermore, adding English training data improved segmentation results, highlighting benefits of multilingual data for low-resource languages. Future work will explore knowledge transfer across languages and extend VSMRC to other underrepresented languages in Asia, Africa or Europe.

\appendix
\section{Prompts for Multiple Choice Data Creation}

Below are the prompts that were used for segment quality validation, multiple-choice question generation, and QA verification as described in Section \ref{sec:multiple_choice_data_creation}. The prompts were designed to work with Vietnamese Wikipedia segments to generate passage-dependent questions.

\subsection{Segment Quality Validation Prompt}
\label{app:validation_prompt}

\begin{small}
\begin{verbatim}
You are given segments from a Vietnamese Wikipedia 
article to review. Your task is to evaluate 
each segment for  its suitability for generating 
educational multiple-choice questions.
Here are the segments:
{batch_xml}
- Evaluate each segment based on:
    1. Sensitive Information: No political
       controversies, religious topics,
       violence, or mature themes.
    2. Accuracy: No misleading, incomplete,
       or culturally insensitive content.
    3. Duplicates: If nearly identical to
       another segment (>80% overlap),
       only accept the most detailed.
    4. Specificity: Contains specific
       entities for unique questions.
    5. Clarity: Clear, unambiguous, and
       free of contradictions.
    6. Distractor Potential: Supports
       questions with plausible but clearly
       wrong distractors.
    7. Complexity: Has 2-5 entities with
       clear connections, not a simple list.
Reply in XML format:
<validation>
<segment id="SEGMENT_ID_1">
    <is_appropriate>yes/no</is_appropriate>
    <reason type="[criterion]">Brief reason
    (max 30 words)</reason>
</segment>
<segment id="SEGMENT_ID_2">
    <is_appropriate>yes/no</is_appropriate>
    <reason type="[criterion]">Brief reason
    (max 30 words)</reason>
</segment>
</validation>
\end{verbatim}
\end{small}

\subsection{Multiple-Choice Question Generation Prompt}
\label{app:generation_prompt}

\begin{small}
\begin{verbatim}
Please analyze the following Vietnamese 
Wikipedia segments and generate one 
multiple-choice question for each segment.
Segments XML:
{segments_xml}
Generate one question for each segment.
Choose from these question types:
  1. Fact-check: Test specific details or
     events within one or two sentences.
  2. Fill-blank: Complete a statement with
     a missing word/phrase.
  3. Reasoning: Synthesize multiple
     sentences to explain a concept.
  4. List: Identify a correct group or
     sequence of items.
General Guidelines:
- Questions and choices must be in
  Vietnamese, clear, and culturally
  appropriate.
- Use 4 choices for all question types.
- Each question must have exactly one
  correct choice.
- Aim for an even distribution across
  question types.
Passage-Dependence Guidelines:
- Questions must be passage-dependent,
  not answerable with general knowledge
  or linguistic patterns.
Distractor Guidelines:
- Distractors must be passage-derived.
- All choices must be of similar length
  and complexity to avoid bias.
Output in XML format:
<qa>
  <segment id="segment_id">
    <question_type>fact_check|
    fill_blank|reasoning|list
    </question_type>
    <question>Question text in Vietnamese
    </question>
    <choices>
      <choice id="0">Choice 1</choice>
      <choice id="1">Choice 2</choice>
      <choice id="2">Choice 3</choice>
      <choice id="3">Choice 4</choice>
    </choices>
    <correct_choice>0|1|2|3</correct_choice>
  </segment>
</qa>
\end{verbatim}
\end{small}

\subsection{QA Verification Prompt}
\label{app:verification_prompt}

\begin{small}
\begin{verbatim}
You are verifying multiple Vietnamese 
multiple-choice questions for an educational
dataset. Review each question, verify its
accuracy, relevance, and format.
For each segment, evaluate:
1. Question clarity, relevance, and passage
   dependence
2. Correct answer accuracy
3. Distractor quality
4. Appropriateness of content
A question is NOT suitable if:
- It contains sensitive political or
  religious topics.
- It is misleading, unclear, ambiguous,
  or contains errors.
- The correct answer is incorrect or
  not supported by the segment.
- There is not exactly one correct answer.
- It can be answered without the segment.
- Distractors are not plausible.
- It is culturally insensitive or
  inappropriate.
Respond with one <verification_results> XML
block containing individual <segment>
elements for each segment ID.
Here are the segments to verify:
{for i, data in enumerate(segments_data, 1):
    Segment: i
    ID: {data['segment_id']})
    Segment Content: {data['segment_text']}
    Question: {data['question_text']}
    Choices:  {data['choices'])} 
    Correct Choice: {data['correct_choice']}
Output Format:
<verification_results>
<segment id="segment_id_1">
    <is_suitable>true/false</is_suitable>
    <reason>Only provide a reason if is_suitable
    is false</reason>
</segment>
...
</verification_results>
\end{verbatim}
\end{small}

\section{Details of types of questions and answers in VSMRC}
\label{app:detail_questions_type}
\begin{table}[b!]
    \centering
    \caption{Examples of the four question types in the VSMRC dataset (Reasoning, Fact-check, List, and Fill-blank) with their Vietnamese text, English translations, and answer choices. Correct answers are shown in bold}
    \label{tab:reasoning_examples}
    \small
    \begin{tabular}{p{2cm}|p{5cm}}
        \hline
        \textbf{Type} & \textbf{Description} \\
        \hline
        Reasoning & {\textbf{Question:} \begin{otherlanguage*}{vietnamese} Ý tưởng cốt yếu trong "thuyết lượng tử cũ" là gì?\end{otherlanguage*} \textit{(What is the essential idea in the "old quantum theory?")} \newline
        A. Tác động vật lý có thể có giá trị bất kỳ. \textit{(Physical action can have any value.)} \newline
        \textbf{B.} Tác động vật lý phải là bội số nguyên của một đại lượng rất nhỏ, "lượng tử của tác động". \textit{(Physical action must be an integer multiple of a very small quantity.)} \newline
        C. Quỹ đạo của hạt không tồn tại. \textit{(Particle trajectories do not exist.)} \newline
        D. Hạt được biểu diễn bằng một hàm sóng. \textit{(Particles are represented by a wave function.)} } \\
        \hline
        Fact-check & {\textbf{Question:} Chiều dài tiêu chuẩn của cá ba sa bằng bao nhiêu lần chiều cao thân? \textit{(The standard length of the Basa fish equals how many times its body height?)} \newline
        A. 2 lần \textit{(2 times)} \newline
        B. 2.5 lần \textit{(2.5 times)} \newline
        C. 3 lần \textit{(3 times)} \newline
        \textbf{D.} 3.5 lần \textit{(3.5 times)} } \\
        \hline
        List & {\textbf{Question:} Các ngành đào tạo đại học tại Trường Đại học Mỹ thuật TP.HCM bao gồm những ngành nào? \textit{(Which undergraduate majors are offered at HCMC University of Fine Arts?)} \newline
        \textbf{A.} Sư phạm mỹ thuật, Hội họa, Đồ họa, Điêu khắc. \textit{(Art pedagogy, Painting, Graphics, Sculpture.)} \newline
        B. Thiết kế thời trang, Kiến trúc, Kỹ thuật. \textit{(Fashion design, Architecture, Engineering.)} \newline
        C. Quản trị kinh doanh, Kế toán, Tài chính. \textit{(Business administration, Accounting, Finance.)} \newline
        D. Luật, Báo chí, Xã hội học. \textit{(Law, Journalism, Sociology.)} } \\
        \hline
        Fill-blank & {\textbf{Question:} Năm \_\_\_\_\_\_ , một trận động đất lớn đã làm sụp đổ toàn bộ phần tường bên ngoài của mặt phía nam của Đấu trường La Mã. \textit{(In the year \_\_\_\_\_\_, a major earthquake caused the collapse of the Colosseum's south face.)} \newline
        A. 100 \newline
        B. 600 \newline
        C. 1200 \newline
        \textbf{D.} 1349 } \\
        \hline
    \end{tabular}
\end{table}

\end{document}